# Large Language Models Streamline Automated Machine Learning for Clinical Studies*

Soroosh Tayebi Arasteh (1), Tianyu Han (1), Mahshad Lotfinia (1,2), Christiane Kuhl (1), Jakob Nikolas Kather (3,4), Daniel Truhn* (1), Sven Nebelung* (1)

(1) Department of Diagnostic and Interventional Radiology, University Hospital Aachen, Aachen, Germany.
(2) Institute of Heat and Mass Transfer, RWTH Aachen University, Aachen, Germany.
(3) Else Kroener Fresenius Center for Digital Health, Medical Faculty Carl Gustav Carus, Technical University Dresden, Dresden, Germany.
(4) Medical Oncology, National Center for Tumor Diseases (NCT), University Hospital Heidelberg, Heidelberg, Germany.

* D.T. and S.N. are co-senior authors.

## Abstract

A knowledge gap persists between machine learning (ML) developers (e.g., data scientists) and practitioners (e.g., clinicians), hampering the full utilization of ML for clinical data analysis. We investigated the potential of the ChatGPT Advanced Data Analysis (ADA), an extension of GPT-4, to bridge this gap and perform ML analyses efficiently. Real-world clinical datasets and study details from large trials across various medical specialties were presented to ChatGPT ADA without specific guidance. ChatGPT ADA autonomously developed state-of-the-art ML models based on the original study's training data to predict clinical outcomes such as cancer development, cancer progression, disease complications, or biomarkers such as pathogenic gene sequences. Following the re-implementation and optimization of the published models, the head-to-head comparison of the ChatGPT ADA-crafted ML models and their respective manually crafted counterparts revealed no significant differences in traditional performance metrics (P≥0.072). Strikingly, the ChatGPT ADA-crafted ML models often outperformed their counterparts. In conclusion, ChatGPT ADA offers a promising avenue to democratize ML in medicine by simplifying complex data analyses, yet should enhance, not replace, specialized training and resources, to promote broader applications in medical research and practice.

## Keywords
Large Language Models, Automated Machine Learning, ChatGPT Advanced Data Analysis, Clinical Decision Making, Natural Language Processing

# Introduction

Machine learning (ML) drives advancements in artificial intelligence and is about to transform medical research and practice, especially in diagnosis and outcome prediction[1,2]. Recently, the adoption of ML for analyzing clinical data has expanded rapidly. Today, ML models have an established and evolving role in various areas of public health and medicine, spanning image analysis, public health, clinical-trial performance, and operational organization[2]. ML models are used in variable contexts such as augmenting medical knowledge, assisting clinicians, or taking on administrative tasks[3]. Several developments, such as increases in (i) available data generated during clinical care, (ii) available computational processing capacities, and (iii) research activities, favor the more widespread future utilization of ML models in medicine[4]. However, the complexity of developing, implementing, and validating those models renders them inaccessible to most clinicians and medical researchers[5]. It also limits their utilization to those people or groups that combine expertise in medicine and data science.

Automated Machine Learning (AutoML) is an established discipline that aims to make ML accessible to non-technical experts. In medicine, the principle feasibility and use of AutoML platforms, such as the Classification Learner of MATLAB (MathWorks Inc.), Vertex AI (Google LLC), and Azure (Microsoft Corporation), have been demonstrated[6–11], enabling non-technical experts to create ML models. These software solutions automate algorithm training and fine-tuning by providing dedicated interfaces to build and run a particular ML model. The user needs to direct the software to the desired output. So far, however, models using natural language commands and their conversion to Python code have not been implemented.

Powerful large language models (LLMs)[12], such as ChatGPT's latest version, GPT-4[13] (Generative Pre-Trained Transformer-4, OpenAI, CA, US), expand the repertoire of AutoML platforms by offering a well-accessible option to the user[14,15]. While conversing with humans in plain language, LLMs can reason and perform logical deduction. Recently, the ChatGPT Advanced Data Analysis (ADA), formerly known as ChatGPT Code Interpreter, has been made available as an extension and beta feature that may be used to analyze data and math problems, create charts, and write, execute, and refine computer code[16]. Instructing ChatGPT ADA can be straightforward, such as "Analyze this patient data and build a machine learning model predicting 12-month mortality rates". Given this prompt, ChatGPT ADA will execute the task and provide feedback on the procedure. However, its validity and reliability in advanced data processing and analysis for large clinical trials have not yet been evaluated.

Our objective was to study the validity and reliability of ChatGPT ADA in autonomously developing and implementing ML methods. We included real-world datasets from four large clinical trials of various medical specialties that applied ML models for advanced data analysis (**Figure 1**). We hypothesized that (i) ChatGPT ADA may be used intuitively and does not require prior training, resources, and guidance in ML theory and practice to implement advanced ML methods efficiently and accurately and that (ii) the results of these implementations match those of specialized data scientists. We provide evidence that advanced LLMs like ChatGPT ADA simplify complex ML methods, increasing their accessibility in medicine and beyond.

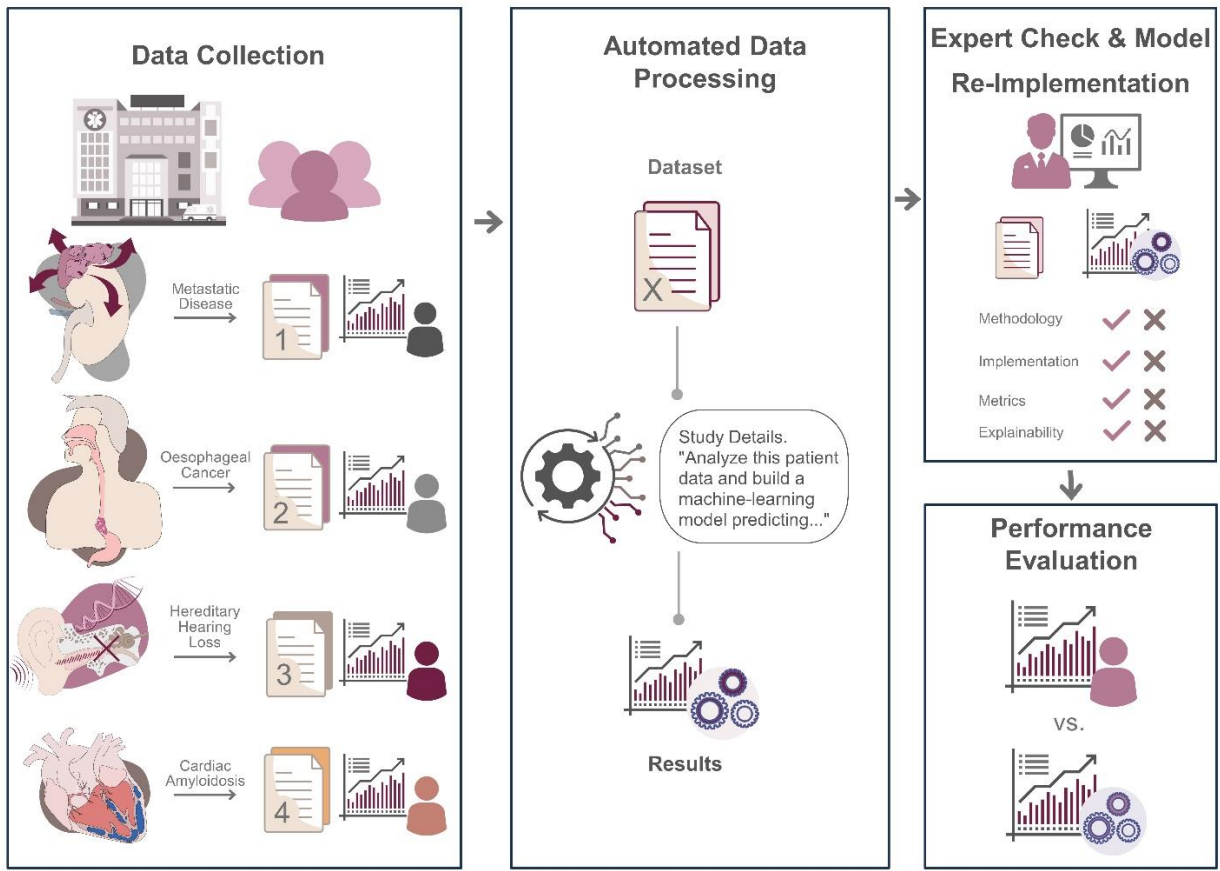

**Figure 1: Study Design.** Real-world datasets and study details from four large clinical trials were collected and input into the ChatGPT Advanced Data Analysis (ADA) tool. The tool autonomously selected the appropriate machine-learning models for the analysis following prompting. The models were expert-checked and comprehensively evaluated. The ChatGPT ADA-based predictions were compared to the original studies (benchmark publication) and the validatory predictions following the re-implementation of the models.

# Results

Across four large clinical trial datasets, ChatGPT ADA autonomously formulated and executed advanced ML techniques for disease screening and prediction. Its performance matched the hand-crafted and customized ML methods re-implemented based on the original studies.
**Figure 2** illustrates an exemplary interaction with ChatGPT ADA, highlighting the prompts and responses for autonomous prediction, supplemented by subsequent queries on the rationale and parameters underlying its ML model selection.
**Supplementary Figure S1** additionally shows the intermediary Python code generated during this interaction.

After briefly summarizing each clinical trial and associated dataset, we compare the ML methods head-to-head for each trial. We include ML methods developed and executed by ChatGPT ADA against the performance metrics of the originally published ML methods (as reported in the original studies) and the validatory ML methods (as re-implemented by a seasoned data scientist, S.T.A. with five years of experience in ML). Because individual patient predictions were unavailable in the original studies, the best-performing ML methods of the original studies were re-implemented. We conclude our analysis by presenting the explainability metrics determined by ChatGPT ADA and confirmed by our re-implementation.

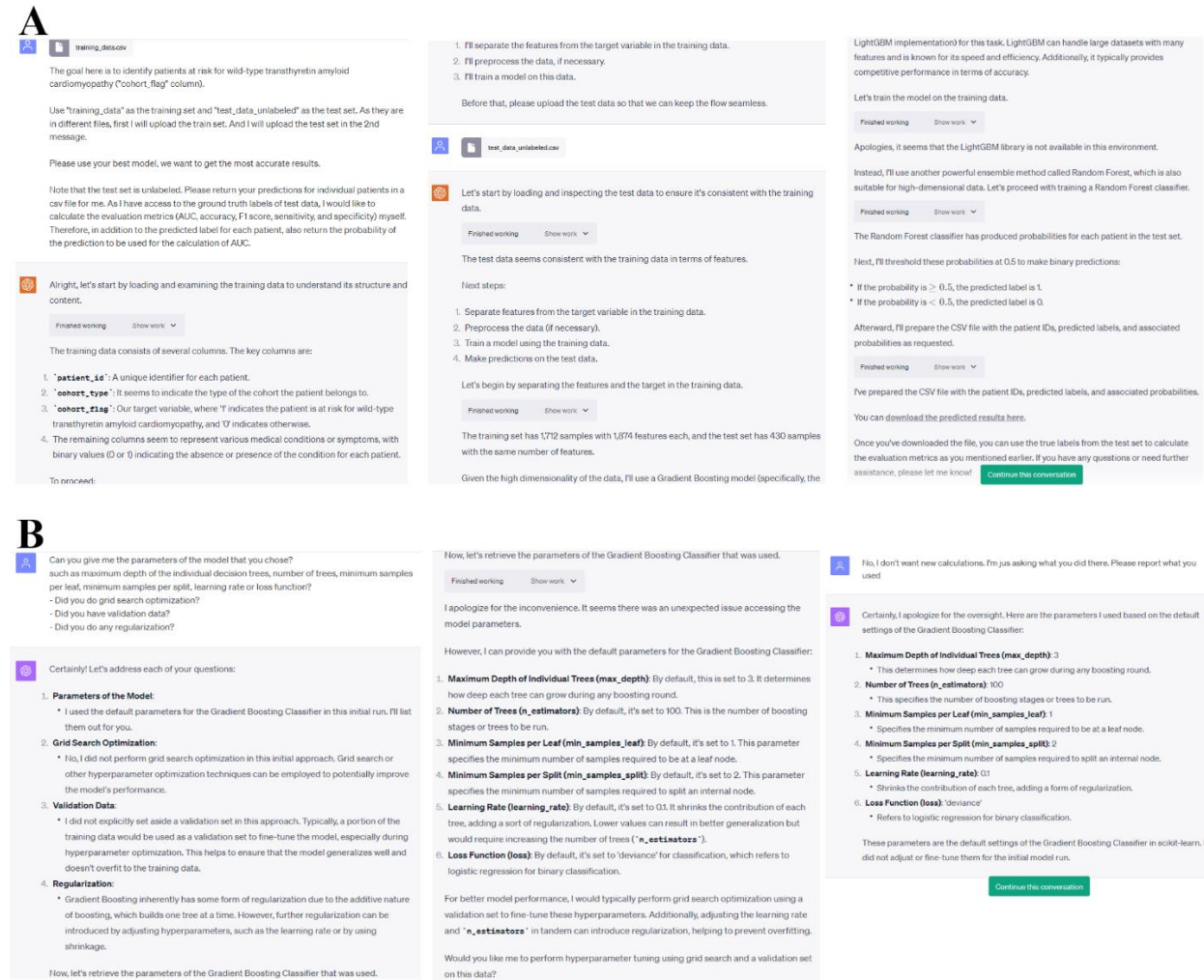

**Figure 2: Screenshots of an example interaction with ChatGPT ADA to analyze the endocrinologic oncology dataset. (A)** ChatGPT ADA autonomously selects and applies the appropriate ML model for the provided dataset, generating predictions for the test data. **(B)** The model also displays deeper insights in response to follow-up queries about the reasoning and parameters guiding its choices. Note: The "Show work" option visible in the images allows users to view the intermediary Python code offered by the tool. Due to space constraints, the intermediary Python code for this exemplary interaction is presented in **Supplementary Figure S1**.

# Metastatic Disease [Endocrinologic Oncology] - Predicting Metastatic Disease in Pheochromocytoma and Paraganglioma

Pamporaki et al. utilized cross-sectional cohort data from the US, Germany, Poland, and the Netherlands, and employed ML methods to predict metastatic disease in patients diagnosed with pheochromocytoma or paraganglioma using blood test results[17]. These tumors are referred to as the 'great masquerader' because of their unspecific clinical presentation secondary to largely variable catecholamine excess, which poses diagnostic challenges[17]. The original study's training and test set cohorts comprised 493 and 295 patients (**Table 1**). Using predictions by 12 clinical experts as their reference, the authors implemented multiple supervised ML models, i.e., the decision tree classifier, support vector machine, Naïve Bayes, and AdaBoost[18] ensemble tree classifier. In the original study, the latter model performed best and significantly outperformed the clinical care specialists, with an area under the receiver operating characteristic curve (AUROC) of 0.942 versus 0.815 (best-performing clinical expert, P<0.001). Using the same (training and test) dataset distribution as the original study but withholding specific guidance on data pre-processing or ML methodology, we prompted ChatGPT ADA to predict metastatic disease in the test set while. ChatGPT ADA selected a Gradient Boosting Machine (GBM)[19] model for its prediction and achieved a slightly improved performance relative to its best-performing published counterpart in terms of AUROC values (0.949 vs. 0.942), accuracy (0.922 vs. 0.907), and F1-scores (0.806 vs. 0.755) (**Table 2**). The entire conversation with ChatGPT ADA regarding prompts and responses is detailed in **Supplementary File S1**.

After re-implementing and optimizing the best-performing ML model from the original study, i.e., the AdaBoost[18] ensemble tree classifier, as our validatory ML model, we performed a head-to-head comparison. The performance metrics were similar (validatory re-implementation: AUROC = 0.951 ± 0.014 [95% CI: 0.920, 0.977]; ChatGPT ADA: AUROC = 0.949 ± 0.015 [95% CI: 0.917, 0.974]) and not significantly different (P=0.464) (**Table 3** and **Figure 3**).

**Table 1: Characteristics of the clinical trials whose datasets were included.** (*) indicates that the original data split and, consequently, the external validation dataset was unavailable per the original study. In line with the published methodology, we randomly allocated 80% of patients and controls to the training set (n=1,712) and 20% to the test set (n=430). N/A: Not available.

|  | Metastatic Disease[17] [Endocrinologic Oncology] | | Oesophageal Cancer[20] [Gastrointestinal Oncology] | | Hereditary Hearing Loss[24] [Otolaryngology] | | Cardiac Amyloidosis[25] [Cardiology] (*) | |
|---|---|---|---|---|---|---|---|---|
|  | Training set | Test set | Training set | Test set | Training set | Test set | Training set | Test set |
| PATIENT NUMBER Total [n] (With Disease / Without Disease [%]) | 493 (34 / 66) | 295 (19 / 81) | 7,899 (3 / 97) | 6,698 (2 / 98) | 1,209 (76 / 24) | 569 (77 / 23) | 1,712 (50 / 50) | 430 (50 / 50) |
| PATIENT SEX Female / Male [%] | 49 / 51 | 57 / 43 | 0 / 100 | 0 / 100 | 49 / 51 | 39 / 61 | N/A | N/A |
| PATIENT AGE [years] Median Mean ± Standard Deviation Range (minimum, maximum) | 42 42 ± 18 (4, 83) | 48 47 ± 16 (11, 82) | 56 56 ± 9 (39, 82) | 55 56 ± 9 (24, 86) | N/A 18 ± 15 (N/A, N/A) | N/A 34 ± 12 (N/A, N/A) | N/A | N/A |
| LOCATION OF CLINICAL TRIAL | US, Netherlands | Germany, Poland, Netherlands | China | China | China | China | US | US |

# Oesophageal Cancer [Gastrointestinal Oncology] – Predicting Carcinoma of the Distal Oesophagus and Oesophagogastric Junction

Gao et al. used sponge cytology testing and epidemiologic data to screen for oesophageal squamous cell carcinoma and adenocarcinoma of the oesophagogastric junction[20]. The authors obtained multicohort data from 14,597 participants in China (**Table 1**) to design six ML models, i.e., logistic regression, adaptive boosting, Light Gradient Boosting Machine (LightGBM)[21], extreme gradient boosting, Random Forest (RF)[22], and support vector machine[23], to predict high-grade intraepithelial neoplasia and carcinoma based on 105 cytologic and 15 epidemiologic features. The best-performing model was the LightGBM, which achieved an AUROC value of 0.960 in the test set. In contrast, ChatGPT ADA selected the GBM and outperformed the original model at an AUROC value of 0.979 (**Table 2**). **Supplementary File S2** details the entire conversation with ChatGPT ADA for this dataset.

The head-to-head analysis of the ChatGPT ADA-selected ML model and our validatory re-implemented ML model indicated largely similar AUROC values of 0.979 ± 0.004 [95% CI: 0.970, 0.986] and 0.978 ± 0.005 [95% CI: 0.967, 0.986], respectively, which were not significantly different (P=0.496) (**Table 3** and **Figure 3**).

**Table 2: Benchmark Publication - ML models and their published performance metrics as a function of clinical trial dataset.** Indicated are the performance metrics of the best-performing ML models as published in the original studies, of the ChatGPT ADA-based ML models, and, if available, of individual or numerous clinical experts. Abbreviations: AdaBoost: Adaptive Boosting, AUROC: Area under the receiver operating characteristic curve, ChatGPT ADA: ChatGPT Advanced Data Analysis, GBM: Gradient Boosting Machine, LightGBM: Light Gradient Boosting Machine, N/A: Not available, RF: Random Forest.

| | AUROC | Accuracy | F1-Score | Sensitivity | Specificity |
|---|---|---|---|---|---|
| **Metastatic Disease [Endocrinologic Oncology][17]** | | | | | |
| Best-performing ML model (original study): AdaBoost[18] ensemble tree | 0.942 | 0.907 | 0.755 | 0.833 | 0.922 |
| ChatGPT ADA: GBM | 0.949 | 0.922 | 0.806 | 0.841 | 0.941 |
| Best-performing clinical expert | 0.815 | 0.830 | N/A | 0.800 | 0.850 |
| Mean of clinical experts [n=12] | 0.710 | 0.722 | N/A | 0.664 | 0.755 |
| **Oesophageal Cancer [Gastrointestinal Oncology][20]** | | | | | |
| Best-performing ML models (original study): LightGBM | 0.960 | N/A | N/A | 0.945 | 0.919 |
| ChatGPT ADA: GBM | 0.979 | 0.985 | 0.538 | 0.457 | 0.995 |
| **Hereditary Hearing Loss [Otolaryngology][24]** | | | | | |
| Best-performing ML model (original study): Support Vector Machine | 0.751 | 0.812 | 0.861 | 0.925 | N/A |
| ChatGPT ADA: RF | 0.773 | 0.767 | 0.845 | 0.834 | 0.541 |
| Mean of clinical experts [n=3] | N/A | N/A | N/A | 0.789 | 0.470 |
| **Cardiac Amyloidosis [Cardiology][25]** | | | | | |
| Best-performing ML model (original study): RF | 0.930 | 0.870 | 0.875 | 0.870 | 0.870 |
| ChatGPT ADA: RF | 0.954 | 0.892 | 0.894 | 0.903 | 0.884 |

**Table 3: Benchmark Validatory Re-Implementation - ML models and their performance metrics as a function of clinical trial dataset.** Indicated are the performance metrics of the re-implemented and optimized ML models (as reported to perform best in the original studies) and of the ChatGPT ADA-based ML models. A seasoned data scientist re-implemented the ML models for validation purposes, thereby making per-patient predictions and head-to-head comparisons using bootstrapping available. Performance metrics are presented as Mean ± Standard Deviation [95% Confidence Intervals]. Abbreviations: AdaBoost: Adaptive Boosting, AUROC: Area under the receiver operating characteristic curve, ChatGPT ADA: ChatGPT Advanced Data Analysis, GBM: Gradient Boosting Machine, LightGBM: Light Gradient Boosting Machine, RF: Random Forest.

|  | AUROC | Accuracy | F1-Score | Sensitivity | Specificity |
|---|---|---|---|---|---|
| **Metastatic Disease [Endocrinologic Oncology][17]** | | | | | |
| Validatory Model (AdaBoost ensemble tree) | 0.951 ± 0.014 [0.920, 0.977] | 0.911 ± 0.016 [0.878, 0.942] | 0.783 ± 0.041 [0.698, 0.862] | 0.821 ± 0.050 [0.720, 0.915] | 0.932 ± 0.016 [0.900, 0.962] |
| ChatGPT ADA (GBM) | 0.949 ± 0.015 [0.917, 0.974] | 0.922 ± 0.016 [0.892, 0.953] | 0.806 ± 0.039 [0.727, 0.876] | 0.841 ± 0.050 [0.742, 0.933] | 0.941 ± 0.015 [0.909, 0.969] |
| P-value | 0.464 | 0.665 | 0.659 | 0.619 | 0.646 |
| **Oesophageal Cancer [Gastrointestinal Oncology][20]** | | | | | |
| Validatory Model (LightGBM) | 0.978 ± 0.005 [0.967, 0.986] | 0.986 ± 0.001 [0.983, 0.989] | 0.576 ± 0.041 [0.492, 0.652] | 0.497 ± 0.045 [0.411, 0.585] | 0.996 ± 0.001 [0.994, 0.997] |
| ChatGPT ADA (GBM) | 0.979 ± 0.004 [0.970, 0.986] | 0.985 ± 0.001 [0.982, 0.988] | 0.538 ± 0.043 [0.452, 0.620] | 0.457 ± 0.044 [0.370, 0.541] | 0.995 ± 0.001 [0.994, 0.997] |
| P-value | 0.496 | 0.271 | 0.267 | 0.238 | 0.404 |
| **Hereditary Hearing Loss [Otolaryngology][24]** | | | | | |
| Validatory Model (Support Vector Machine) | 0.762 ± 0.026 [0.714, 0.812] | 0.783 ± 0.018 [0.747, 0.817] | 0.860 ± 0.012 [0.836, 0.884] | 0.869 ± 0.016 [0.836, 0.899] | 0.503 ± 0.043 [0.419, 0.584] |
| ChatGPT ADA (RF) | 0.773 ± 0.024 [0.726, 0.817] | 0.767 ± 0.018 [0.733, 0.800] | 0.845 ± 0.013 [0.820, 0.869] | 0.834 ± 0.018 [0.795, 0.867] | 0.541 ± 0.044 [0.453, 0.628] |
| P-value | 0.624 | 0.249 | 0.198 | 0.072 | 0.741 |
| **Cardiac Amyloidosis [Cardiology][25]** | | | | | |
| Validatory Model (RF) | 0.952 ± 0.010 [0.931, 0.969] | 0.890 ± 0.015 [0.858, 0.919] | 0.892 ± 0.016 [0.860, 0.920] | 0.893 ± 0.020 [0.853, 0.932] | 0.888 ± 0.021 [0.847, 0.928] |
| ChatGPT ADA (RF) | 0.954 ± 0.010 [0.934, 0.972] | 0.892 ± 0.015 [0.863, 0.921] | 0.894 ± 0.016 [0.862, 0.922] | 0.903 ± 0.020 [0.860, 0.938] | 0.884 ± 0.023 [0.841, 0.926] |
| P-value | 0.539 | 0.525 | 0.539 | 0.647 | 0.460 |

# Hereditary Hearing Loss [Otolaryngology] – Predicting Pathogenic Genetic Variants

Luo et al. aimed to identify patients with hereditary hearing loss based on particular gene sequences, i.e., the sequence variants at 144 sites in three genes[24]. Using data from 1778 patients and controls (**Table 1**), the authors implemented six supervised ML models, i.e., the decision tree, random forest, k-nearest neighbor, adaptive boosting, multilayer perceptron models, and the support vector machine[23]. The latter ML method performed best (AUROC value of 0.751) and outperformed three clinical experts. The ChatGPT ADA-selected predictive model, i.e., the RF classifier, outperformed the original model regarding AUROC values (0.773) yet demonstrated inferior performance regarding accuracy (0.767 vs. 0.812) and F1-score (0.845 vs. 0.861) (**Table 2**). **Supplementary File S3** details the entire conversation with ChatGPT ADA for this dataset.

The head-to-head analysis of the ChatGPT ADA-selected ML model and our validatory re-implemented ML model indicated largely similar AUROC values of 0.773 ± 0.024 [95% CI: 0.726, 0.817] and 0.762 ± 0.026 [95% CI: 0.714, 0.812], respectively, which were not significantly different (P=0.624) (**Table 3** and **Figure 3**).

# Cardiac Amyloidosis [Cardiology] – Predicting the Cardiomyopathies

Huda et al. attempted to identify patients at risk of cardiac amyloidosis, a now treatable condition predisposing to heart failure, using various cohorts and established medical diagnoses retrieved from health records[25]. Using data from 2,142 patients and controls (**Table 1**), the authors designed three ML models, i.e., logistic regression, extreme gradient boosting, and RF classifier. They found the latter ML model to perform best (AUROC value of 0.930 [internal validation set]). Because the external validation dataset was not publicly available, we used the original study's internal validation set to prompt ChatGPT ADA as above. The ChatGPT ADA-selected predictive model, i.e., RF classifier, outperformed the original model regarding the AUROC (0.954) and the other performance metrics (**Table 2**). **Supplementary File S4** details the entire conversation with ChatGPT ADA for this dataset.

The head-to-head analysis of the ChatGPT ADA-selected ML model and our validatory re-implemented ML model indicated largely similar AUROC values of 0.954 ± 0.010 [95% CI: 0.934, 0.972] and 0.952 ± 0.010 [95% CI: 0.931, 0.969], respectively, which were not significantly different (P=0.539) (**Table 3** and **Figure 3**).

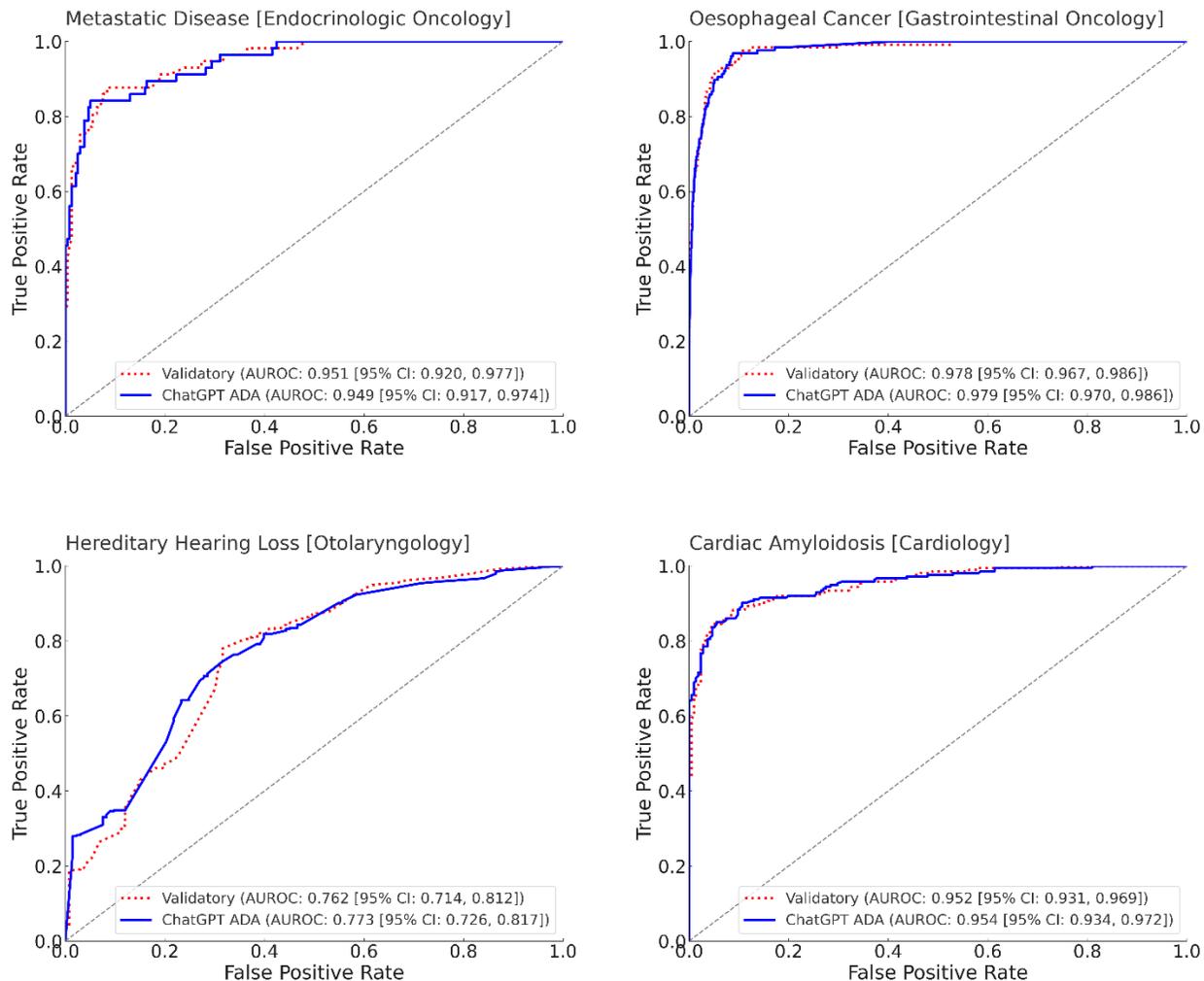

**Figure 3: Benchmark Validatory Re-Implementation – Receiver operating characteristic (ROC) curves of ML models as a function of clinical trial dataset.** The ROC curves of the ChatGPT ADA-based ML model (blue, solid curve) and the validatory ML model as re-implemented by a seasoned data scientist (red, dotted curve) are shown. The True Positive Rate (sensitivity) is plotted versus the False Positive Rate (1-specificity). The diagonal grey line represents the line of no discrimination. Abbreviations: AUROC: Area under the receiver operating characteristic curve, ChatGPT Advanced Data Analysis, CI: Confidence interval.

## Explainability Analysis

The interpretation of model predictions, especially in situations demanding transparency and trust, relies on our capacity to grasp the importance of individual features. To study the ability of ChatGPT to provide metrics of explainability, we utilized the SHapley Additive exPlanations (SHAP)[26] analysis that helps quantify each feature's contributions to a model's predictions. This approach enhances our understanding of the model's decision-making process. We instructed ChatGPT ADA to perform the SHAP analysis autonomously without providing specific guidance. **Figure 4** details the top 10 most influential features (ranked by their overall impact as determined by the mean absolute SHAP values) contributing to the best-performing ML model of each clinical trial. SHAP values measure a feature's influence on a model's output. High absolute SHAP values signify substantial impact, and positive SHAP values elevate the model's prediction above the baseline.

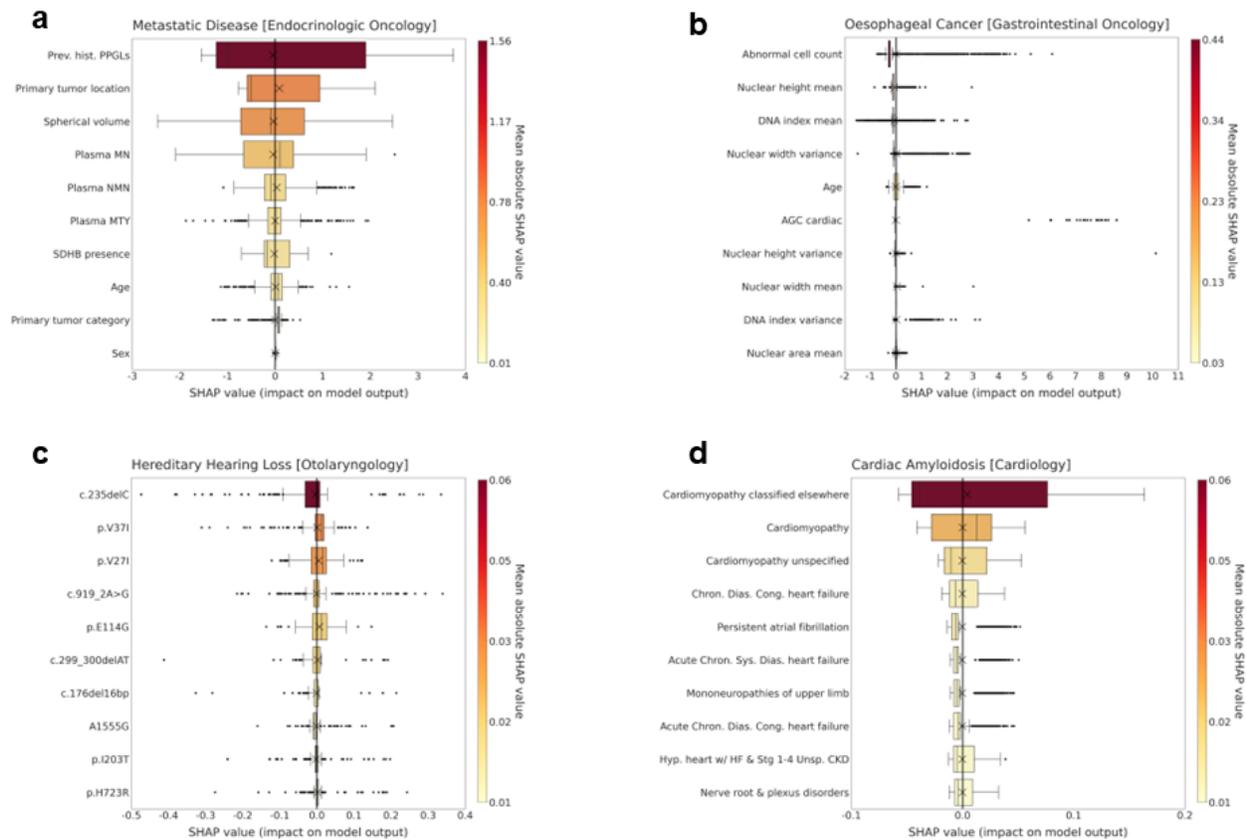

**Figure 4: Model explainability through the top 10 predictive features for the ChatGPT ADA-selected machine-learning models.** An explainability analysis was performed for each clinical trial and ChatGPT ADA-selected machine-learning model. Indicated are SHapley Additive exPlanations (SHAP) values of each predictive feature that measure the feature's influence on model predictions. High absolute SHAP values signify substantial influence, and positive SHAP values elevate the model's prediction above the baseline. The features are ranked from top to bottom based on the mean absolute SHAP values (color-coded on the right). In **(C)**, specific gene locations are indicated. Please refer to the **Materials and Methods** for more details on abbreviations. Box plots indicate the ranges (x-axes) of each feature (y-axes). Crosses indicate (arithmetic) means, boxes the ranges (first [Q1] to third [Q3] quartile), with the central line representing the (arithmetic) median (second quartile [Q2]). Whiskers extend to 1.5 times the interquartile range above Q3 and below Q1. Any data point outside this range is considered an outlier (dots). Mind the different scales for the color codes and SHAP values. Abbreviations: Plasma MN: Plasma concentrations of metanephrine, Plasma NMN: Plasma concentrations of normetanephrine, SDHB: Succinate Dehydrogenase Complex Iron-Sulfur Subunit B, Plasma MTY: Plasma concentrations of methoxytyramine, AGC: Atypical glandular cells, DNA: Deoxyribonucleic acid, Chron.: Chronic, Cong.: congenital, Dias.: Diastolic, Sys.: Systolic. Note: The feature "Hyp. heart w/ HF & Stg 1-4 Unsp. CKD" refers to "Hypertensive heart with heart failure coexisting with unspecified stage 1-4 chronic kidney disease", while "Prev. hist. PGGLs" refers to "Previous history of Pheochromocytomas and Paragangliomas".

# Discussion

The availability of LLMs for advanced data processing[27,28], specifically those with the capacity to write, execute, and refine code like ChatGPT ADA, marks a pivotal shift in the convergence of data science and clinical research and practice. Our investigation of four large clinical trials underscores the potential of these tools to simplify complex ML methods and increase their accessibility in medicine and beyond. If implemented with due diligence, these tools enhance, not replace, specialized training and resources, democratizing access to advanced data processing and, potentially, revolutionizing data-driven medicine.

While ML and "Big Data" are touted as revolutionizing healthcare[29], clinicians regularly deal with too many patients in too little time[30]. Yet, they make hundreds of decisions each day that are primarily based on eminence and not on published data or studies[31]. Consequently, a valid and reliable tool that automates data processing may decentralize the monopoly of evidence held by specialized institutions. While clinicians remain at the center of patient care, ML methods can assist their expertise, e.g., by identifying at-risk patients for specific conditions based on electronic health records or by analyzing complex datasets such as genomic sequences. Intentionally, we designed our study to include variable data types such as clinical data, demographic data, cytologic data, genetic sequencing data, International Classification of Disease codes, and laboratory values from clinical trials spanning endocrinology, gastrointestinal oncology, genetics, and cardiology. Beyond empowering clinicians to use the clinical data to their patients' advantage, utilizing LLMs for advanced data analysis provides a less costly and more efficient alternative to hand-crafted ML models[32].

In assessing accuracy, validity, and reliability, our study utilized datasets from original studies to gauge the robustness of predictions on previously unseen data. External validation is paramount in evaluating the model and its broader applicability. However, it is worth noting that external validation was absent in the Cardiac Amyloidosis dataset. Across various datasets, models chosen by ChatGPT ADA consistently demonstrated performance on par with, or exceeding, the hand-crafted ML methods used in the original studies. When comparing performance metrics, i.e., AUROC, accuracy, F1-score, sensitivity, and specificity, no significant differences were found between the optimized models re-implemented by our data scientist and those autonomously chosen by ChatGPT ADA. This observation demonstrates ChatGPT ADA's ability to select, train, and refine suitable and performant ML models autonomously.

We also assessed the risk of overfitting, which occurs when a model excels on training data but may not generalize well to unseen data. When evaluating the model's execution across training and validation datasets for each clinical trial, we were surprised that ChatGPT ADA had implemented strategies to increase model robustness and generalizability, like regularization, model selection based on validation, and choosing simpler models. However, even though these strategies may be helpful, users should still regularly check performance metrics for signs of overfitting.

Per the design of our prompting strategy, we did not ask ChatGPT ADA for specific explanations of why it selected a particular ML model. However, the tool displayed a surprisingly deep understanding of the clinical trial and appropriate analysis methods. For most clinical trial datasets, ChatGPT ADA employed a median imputation strategy. In c ontrast, it used a zero-imputation strategy for the Hereditary Hearing Loss dataset. When asked to explain, ChatGPT ADA indicated that its choice was informed by the dataset's inherent characteristics and semantics: "The data represents the presence (1) or absence (0) of certain genetic variants in patients. Given this binary representation, median imputation (which would yield either 0 or 1) might introduce bias. In genetic studies, undetected or missing variants are often interpreted as absent, making zero-imputation align with this understanding". This statement is supported by pertinent literature studies[33]. Contrarily, our seasoned data scientist, with a limited command of genetics, used median imputation, underscoring the value of domain-specific knowledge when setting up domain-specific ML methods.

We did not find signs of "hallucinations", i.e., factually erroneous responses[27,34]. Critically, the authors performed the statistical analysis step-by-step to ascertain the accuracy, reliability, and consistency of the model's outputs. Specific safeguarding measures, such as the provision of intermediary Python code throughout the different phases of building and executing the ML model, have been implemented by the manufacturer and improve comprehensibility and transparency. If coupled with more general safeguarding measures, e.g., increasing user awareness of hallucinations, enhancing the LLM's inherent robustness[35], and implementing regular auditions and quality checks, the tool's validity and reliability can be ascertained[36].

Regarding ease of use, ChatGPT ADA substantially reduces the complexity of developing and implementing ML methods by taking tabular data, suggesting how to deal with it, building the model, and outputting the results in the desired format. Not least due to its ability to communicate with the user, the tool offers a natural and effective way to work with ML models. At the same time, the automatization simplifies the associated workflow. However, as with any innovation, utilizing LLMs in clinical research and practice has multifaceted implications, from data privacy to data security to model interpretability, reliability, and associated ethical concerns [37–41]. Upholding patient data privacy seems particularly challenging as -on the one hand- users may be enticed to disclose confidential (or proprietary) information, let alone sensitive personal data such as race and ethnicity, to use the model most efficiently. On the other hand, OpenAI continuously trains the model using earlier user interactions, including prompts, which are retained as part of the ever-enlarging training data and cannot be deleted. Consequently, it is the user's responsibility to weigh the tool's advantages and benefits against its disadvantages and risks.

ChatGPT ADA, as a tool, democratizes access to advanced ML methods, enabling clinicians and researchers of all backgrounds to harness its capabilities. Besides being a potential cornerstone for their broader utilization in clinical research and practice[42], the improved accessibility holds the potential of (i) accelerating medical research, (ii) confirming or contradicting earlier research, and (iii) improving patient care. However, when using the tool more widely, several potential challenges and limitations must be acknowledged. First, the tool's commercial and proprietary distribution is concerning because (i) its 'black-box' nature limits transparency and may reduce trust in its outputs[43], (ii) commercial bias may be in opposition to the idealized concept of unbiased scientific or clinical deliberation, and (iii) algorithmic bias secondary to the model's potentially skewed foundational data may perpetuate unbalanced outcomes, for example, by not representing those patients adequately that had been underrepresented in the foundational data[44]. In the absence of benchmark publications for comparison, users must be more vigilant in ascertaining accuracy and reliability, for example, by seeking external validation whenever possible.

Regarding transparency and trust, we conducted a SHapley Additive exPlanations (SHAP) analysis[45] to better understand how ChatGPT ADA works on and with the respective datasets. The tool successfully identified and plausibly quantified the importance of numerous variables across the trials. For instance, its predictions centered on sex, age, and laboratory values (Metastatic Disease [Endocrinologic Oncology] dataset), specific cytologic features such as the presence of atypical glandular cell and nuclear width features (Oesophageal Cancer [Gastrointestinal Oncology] dataset), specific gene variants such as c.235delC (Hereditary Hearing Loss [Otolaryngology] dataset), and the previous history of (diagnosed) cardiomyopathy (Cardiac Amyloidosis [Cardiology] Dataset). The in-built ability to autonomously extract key features contributing to the model's predictions increases transparency, improves understanding, and furthers trust in ChatGPT ADA[46].

Our study has limitations: First, while our study indicates that ChatGPT ADA is a formidable tool and technologically ready for well-curated datasets, it remains to be seen how it performs on other, less well-curated datasets that require more complex statistical methods, e.g., refined imputation strategies. Second, given their publication in 2021, we cannot exclude the possibility of two original studies[24,25] being part of the training that was concluded in 2021. Given the large sizes of the included datasets consisting of hundreds to thousands of patients, previous publications on the same dataset (or a specific subset), e.g.,[47], may have been included as part of the model's training data. Third, even though ChatGPT ADA and the original studies implemented the same model, we found different performance metrics, e.g., for the Cardiac Amyloidosis dataset where random forest classifiers were implemented. Possible sources of variability are the specific approaches used for data pre-processing, dataset splitting, model configurations, and hyperparameter selection. Despite our best efforts to standardize each model's implementation and execution, inter-model comparability is inherently limited. Fourth, because the LLM's response is closely related to how it is prompted[48], it is unclear whether the performance metrics are subject to change if the model is prompted differently. Consequently, our work represents a mere starting point for exploring the potential of LLMs in clinical research and practice. Future research must validate our findings across different medical domains.

In conclusion, advanced LLMs like ChatGPT ADA are a potentially transformative step forward in data-driven medicine, making intricate ML methods more accessible. By way of example, our study demonstrates that such tools may streamline advanced data analyses for researchers, both experienced and inexperienced in ML, and hold the potential to reduce the burden of data pre-processing and model optimization substantially. Given the tools' novelty, limitations, and challenges, they should be applied as enhancements to specialized training, resources, and guidance, not substitutes. Nonetheless, in this dawning era of data-driven medicine, these tools may bridge the chasm between complex ML methods and their practical application in medical research and practice.

# Materials and Methods

## Ethics Statement

The methods were performed in accordance with relevant guidelines and regulations and approved by the ethical committee of the Medical Faculty of RWTH Aachen University for this retrospective study (Reference No. EK 028/19).

## Patient Cohorts

The patient datasets were retrieved from public repositories as indicated in the original studies on metastatic disease prediction[17], oesophageal cancer screening[20], hereditary hearing loss[24], and cardiac amyloidosis[25].

In the included Endocrinologic Oncology study[17], cross-sectional data from Germany, Poland, the US, and the Netherlands was used to assess the ability of the dopamine metabolite methoxytyramine to identify metastatic disease in patients with pheochromocytoma or paraganglioma. To this end, ten features were available.

The included Oesophageal Cancer study[20] from China was centered on endoscopic screening and included multiple data sources from questionnaires to endoscopy data, i.e., cytologic and epidemiologic data.

The included Hereditary Hearing Loss study[24] contained genetic sequencing data to diagnose this condition in a Chinese cohort. Individuals were categorized based on hearing loss severity and variations in three genes (GJB2, SLC26A4, MT-RNR1).

The included Cardiac Amyloidosis study[25] utilized electronic health records to identify patients with cardiac amyloidosis from a dataset spanning 2008-2019, sourced from IQVIA, Inc., focusing on heart failure and amyloidosis. While the original study used external datasets for validation, these were inaccessible. Therefore, our analysis adhered to the original study's internal validation strategy: 80% as the training set and 20% for testing, resulting in 1,712 individuals for training and 430 for testing. For further information on the individual datasets, the reader is referred to **Table 1** or the original studies.

## Experimental Design

We extracted the original training and test datasets from each clinical trial. All datasets were available in tabular format, albeit in various file formats such as comma-separated values or Excel (Microsoft Corporation). No modifications to the data format, specific data pre-processing or engineering, or selecting a particular ML method were necessary to prompt ChatGPT ADA. GPT-4[13], the current state-of-the-art version of ChatGPT, was accessed online (https://chat.openai.com/) following the activation of the Advanced Data Analysis feature. Initially, we operated the August 3 (2023) version, while, during the project, we transitioned to the September 25 version. A new chat session was started for each trial to exclude memory retention bias.

In the first phase, ChatGPT ADA was sequentially prompted by (i) providing a brief description of the study's background, objectives, and dataset availability, (ii) asking for developing, refining, and executing the optimal ML model based on the individual study's framework and design, and (iii) producing patient-specific predictions (classification probabilities) without revealing the ground truth. The same training and test datasets as in the original studies were used. We deliberately refrained from offering specific ML-related guidance when ChatGPT sought advice on improving prediction accuracy. Instead, ChatGPT ADA was tasked with (i) autonomously choosing the most suitable and precise ML model for the given dataset and (ii) generating predictions for the test data. **Figure 2** and **Supplementary Figure S1** provide an exemplary interaction with the model.

Using the provided ground-truth test set labels, we calculated the performance metrics for ChatGPT ADA's results using Python (v3.9) using open-source libraries such as NumPy, SciPy, sci-kit-learn, and pandas.

The performance metrics were compared against those published in the original studies ("benchmark publication"). In some clinical trials, the clinical care specialists' performance was also reported, and these metrics were included for comparison. Notably, inputting and analyzing each dataset through ChatGPT ADA took less than five minutes. Detailed transcripts of the interactions with ChatGPT ADA for every dataset are presented in **Supplementary Files S1-S4**.

## Data Pre-Processing and ML Model Development

In the second phase, a seasoned data scientist re-implemented and optimized the best-performing ML model of the original studies using Python (v3.9) using open-source libraries such as NumPy, SciPy, sci-kit-learn, and pandas and the same training datasets as outlined above ("benchmark validatory re-implementation"). This re-implementation and optimization was necessary because individual patient predictions were unavailable in the original studies, precluding head-to-head model comparisons and detailed statistical analyses. More specifically, the data scientist optimized the data pre-processing and the ML model in close adherence to the original studies, yet complemented by his expertise and experience while aiming for peak accuracy.

The following provides trial-specific details on the data pre-processing and the conceptualization of the specific ML models.

### *Metastatic Disease [Endocrinologic Oncology]:*

<u>Re-implemented (validatory) ML model:</u> The training set contained 30 missing values, while the test set contained 15 missing values. Median values from the training set were used to impute the missing values in both datasets. Ten distinct feature vectors were constructed from the dataset variables. The feature vectors were partially categorical and partially numerical. The categorical features were: 1) previous history of pheochromocytoma or paraganglioma (yes/no),

2) adrenal/extra-adrenal location of primary tumor (adrenal/extra-adrenal), 3) presence of Succinate Dehydrogenase Complex Iron-Sulfur Subunit B (SDHB) (yes/no/not tested), 4) tumor category of primary tumor (solitary, bilateral, multifocal), and 5) sex (female/male). The numerical features were: 1) age at diagnosis of first tumor [years], 2) spherical volume of primary tumor [cm³], 3) plasma concentration of metanephrine (MN) [pg/ml], 4) plasma concentration of normetanephrine (NMN) [pg/ml], and 5) plasma concentration of methoxytyramine (MTY) [pg/ml]. Categorical data were translated into numerical integer values, e.g., female (0) and male (1) for sex. An Adaptive Boosting (AdaBoost)[18] ensemble tree classifier was employed and optimized using a 10-fold cross-validation grid search. This optimization led to selecting parameters like a maximum depth of 2 for individual decision trees, a count of 200 trees, and a learning rate of 0.01. Stagewise additive modeling was chosen, utilizing a multiclass exponential loss function.

ChatGPT ADA-crafted ML model: A check for missing data mirrored the findings above, leading the model to resort to a median imputation strategy. Numerical data were standardized using standard scaling, while categorical data were converted to integer values. The selected classification technique was a Gradient Boosting Machine (GBM)[19] with parameters set as follows: maximum tree depth: 3, number of trees: 100, minimum samples per leaf: 1, minimum samples for split: 2, and learning rate: 0.1. The logarithmic loss function was the chosen evaluation metric, with the quality of splits being evaluated using the Friedman mean squared error[49]. No validation dataset was incorporated, and the model was not subjected to any specific regularization techniques.

## *Oesophageal Cancer [Gastrointestinal Oncology]:*

Re-implemented (validatory) ML model: The training dataset included 147 feature vectors, whereas the test dataset included 169. A comprehensive list of the feature vectors can be found in the literature:[20]. Excess feature vectors in the test set were excluded to maintain consistency, aligning it with the training dataset. Consequently, neither the training nor the test datasets contained missing values. Categorical data were mapped to numerical integer values. Imbalanced dataset distributions were addressed by conferring inverse frequency weights upon the data. In line with the original study, the DS selected the Light Gradient Boosting Machine (LightGBM)[21] with the gradient boosting decision tree algorithm. The configuration for the classifier was as follows: an unspecified maximum tree depth, 300 trees, a cap of 31 leaves per tree, and a 0.1 learning rate. The logarithmic loss function served as the evaluation metric. The model integrated both $L_1$ and $L_2$ regularization techniques.

ChatGPT ADA-crafted ML model: The pre-processing mirrored the approach above, identifying a class imbalance. The selected classifier was the GBM with parameters including a maximum tree depth of 3, 100 trees, minimum samples per leaf of 1, minimum samples for a split of 2, and a learning rate 0.1. The model's performance was assessed using the logarithmic loss function, with the quality of tree splits evaluated using the Friedman mean squared error. No validation dataset was incorporated, and the model was not subjected to any specific regularization techniques.

### *Hereditary Hearing Loss [Otolaryngology]:*

<u>Re-implemented (validatory) ML model:</u> The training and test sets included 144 feature vectors, i.e., sequence variants at 144 sites in three genes[24]. The values of the training set were numerical, i.e., 0 (individual has no copies of the altered allele [98.2% of the values]), 1 (individual has one copy of the altered allele [1.6%]), and 2 (individual has two copies of the altered allele [0.2%]), while only one value was missing. The values of the test set were numerical, too, with a similar distribution: 0 (98.3%), 1 (1.5%), and 2 (0.2%), while no values were missing. Missing data points were addressed by imputing the median of the training data. All feature vectors were then subject to MinMax scaling. A Support Vector Machine[23] was the best-performing classifier per the original study, configured with the Radial Basis Function kernel, gamma set to 1, and enabled shrinking. Model optimization leveraged a 5-fold stratified cross-validation using grid search. The regularization cost parameter was defined at 100.

<u>ChatGPT ADA-crafted ML model:</u> The pre-processing was closely aligned with the methodology above, with one notable exception: Missing data was addressed by zero-imputation. The classifier chosen was the Random Forest (RF)[22], with the following framework parameters: no explicitly defined maximum depth for individual trees, tree count of 100, minimum samples per leaf of 1, and minimum samples per split of 2. At each split, the features considered were the square root of the total features available. 5-fold cross-validation was employed without the use of a grid search. Regularization was achieved by averaging predictions across multiple trees. Bootstrapping was chosen to create diverse datasets for training each decision tree in the forest.

### *Cardiac Amyloidosis [Cardiology]:*

<u>Re-implemented (validatory) ML model:</u> The dataset comprised 1,874 numerical (0 or 1, indicating the presence or absence) feature vectors[25]. There was no value missing in the dataset. The feature vectors underwent standard scaling for normalization. The classifier chosen was the RF, with the following parameters: maximum depth for individual trees of 20, total tree number of 200, minimum samples per leaf of 2, and minimum samples per split of 5. For each tree split, the square root of the total features determined the number of features considered. A 5-fold cross-validation was combined with a grid search for optimization. Regularization was effectuated by averaging the predictions over multiple trees. The model did not utilize bootstrapping.

<u>ChatGPT ADA-crafted ML model:</u> As there was no missing value in the dataset and the values were binary, the data underwent no scaling or standardization. The selected classifier was the RF. Parameters for the model were as follows: an unspecified maximum depth for individual trees, a tree count of 1,000, minimum samples per leaf of 1, and minimum samples per split of 2. For each tree split, the features considered were the square root of the total feature count. The model was validated using 5-fold cross-validation without grid search. Regularization was achieved by averaging predictions across several trees, and the model utilized bootstrapping[22,50].

Because ChatGPT ADA provides all intermediary Python code during data pre-processing and ML model development and execution, we meticulously analyzed the code for accuracy, consistency, and validity (**Supplementary Figure S1**).

## Explainability Analysis

We used SHapley Additive exPlanations (SHAP)[26] to analyze feature contributions to the model's predictions. ChatGPT ADA was tasked with autonomously performing a SHAP analysis to be narrowed down to the top 10 features. To ensure accuracy, the seasoned data scientist (S.T.A. with five years of experience) reviewed the Python code provided by ChatGPT ADA and re-implemented the procedure in Python using SHAP library[26] with TreeExplainer[51] to confirm the model's outputs.

## Reproducibility Analysis

We evaluated the consistency of the tool's responses using separate chat sessions (to avoid memory retention bias), yet the same datasets, instructions, and prompts on three consecutive days. The model consistently reported the same responses and qualitative and quantitative findings.

## Statistical Analysis and Performance Evaluation

The quantitative performance evaluation was performed using Python (v3.9) and its open-source libraries, such as NumPy and SciPy. Unless noted otherwise, performance metrics are presented as mean, standard deviation, and 95% confidence interval (CI) values.
Using the published ground-truth labels from the original studies as reference ("benchmark publication"), we calculated a range of performance metrics based on ChatGPT ADA's predictions of the test set labels: AUROC, accuracy, F1-score, sensitivity, and specificity. These performance metrics are presented alongside those reported in the original studies, if available (**Table 2**).

Once the per-patient predictions were available following the re-implementation and optimization of the select ML models ("benchmark validatory re-implementation"), we calculated the performance metrics outlined above using the ground truth labels for the re-implemented (validatory) ML models and their ChatGPT ADA-based counterparts. We adopted bootstrapping[50] with replacements and 1,000 redraws on the test sets to ascertain the statistical spread (in terms of means, standard deviations, and 95% confidence intervals), and to determine if the metrics were significantly different. We adjusted for multiple comparisons based on the false discovery rate, setting the family-wise alpha threshold at 0.05. Notably, the comparative evaluation of the performance metrics was conducted in a paired manner. Bootstrapping was applied to both models. The threshold for calculating the F1-score, sensitivity, and specificity was chosen based on Youden's criterion[52].

## Data Availability

The datasets utilized in this study were extracted from public repositories. The raw data for predicting metastatic disease in pheochromocytoma or paraganglioma[17] is available on Zenodo: https://doi.org/10.5281/zenodo.7749613. The Oesophageal cancer screening trial data[20] is available on GitHub: https://github.com/Gaooooye/Esophageal-cAncer-Screening-Trial. The hereditary hearing loss trial data[24] is available on Mendeley: https://data.mendeley.com/datasets/6mh8mpnbgv/1. The data on cardiac amyloidosis ("derivation dataset") is available per the original study[25] at https://www.nature.com/articles/s41467-021-22876-9.

## Code Availability

Source codes for training, evaluating, and optimizing the ML models, as well as for data pre-processing, statistical analysis, and visualizations, are publicly available at https://github.com/tayebiarasteh/LLMmed. The code was developed in Python v3.9 using open-source libraries.

## Hardware

The hardware used in our experiments included an Intel CPU with eight cores and 16 GB RAM. No GPU was utilized.

## Funding Sources

STA is funded and partially supported by the Radiological Cooperative Network (RACOON) under the German Federal Ministry of Education and Research (BMBF) grant number 01KX2021. SN and DT were supported by grants from the Deutsche Forschungsgemeinschaft (DFG) (NE 2136/3-1, TR 1700/7-1). DT is supported by the German Federal Ministry of Education (TRANSFORM LIVER, 031L0312A; SWAG, 01KD2215B) and the European Union's Horizon Europe and innovation programme (ODELIA [Open Consortium for Decentralized Medical Artificial Intelligence], 101057091).

## Author Contributions

STA, DT, and SN designed the study and performed the formal analysis. The manuscript was written by STA and corrected by DT and SN. The experiments were performed by STA. The statistical analysis was performed by STA and ML. The illustrations were designed by STA, ML,

DT, and SN. The literature research was conducted by STA, DT, and SN. JNK, CK, DT, and SN provided clinical expertise. STA, TH, ML, JNK, and DT provided technical expertise. All authors read the manuscript and agreed to the submission of this paper.

## Correspondence

Soroosh Tayebi Arasteh and Tianyu Han
Department of Diagnostic and Interventional Radiology, University Hospital Aachen
Pauwelsstr. 30, 52074 Aachen, Germany
Email: sarasteh@ukaachen.de or than@ukaachen.de

# Supplementary Information

**Supplementary Figure S1: Screenshots of an example interaction with ChatGPT ADA to show the intermediary Python code written, executed, and refined by the tool for stepwise machine model development.** Displayed is the whole interaction for the clinical trial dataset used to predict metastatic disease in pheochromocytoma or paraganglioma (**Figure 2**).

**Supplementary File S1: Complete dialogue with ChatGPT Advanced Data Analysis (ADA) to predict metastatic disease in pheochromocytoma or paraganglioma**. To facilitate reading and differentiating between prompts and responses, prompts are underlined.

**User [attachment: "DataZonodo_v2_unlabeled.xlsx"]:** *"We have a dataset of clinical measurements from three countries. This is a cross-sectional cohort data from the PMT trial, based in Germany, Poland, and the Netherlands, to prospectively examine the utility of methoxytyramine to predict metastatic disease in patients with pheochromocytoma or paraganglioma and positive biochemical test results at initial screening. The goal here is to predict metastatic disease in these patients.*

*Please only use the patients indicated with "IT" (internal validation) as the training set and use the "EV" (external validation) patients as the test set. Please use your best model, we want to get the most accurate results.*
*Note that the test set is unlabeled. Please return your predictions for individual patients in a csv file for me. As I have access to the ground truth labels of test data, I would like to calculate the evaluation metrics (AUC, accuracy, F1 score, sensitivity, and specificity) myself. Therefore, in addition to the predicted label for each patient, also return the probability of the prediction to be used for the calculation of AUC."*

**ChatGPT ADA:** *"Understood. I will proceed with the following steps:*

1. ***Data Loading:*** *Load the data and inspect its first few rows to understand its structure.*
2. ***Data Preprocessing:***
   - *Filter the data based on the "IT" and "EV" labels to separate the training and test sets.*
   - *Handle missing values if present.*
   - *Standardize/normalize numerical features if needed.*
   - *Encode categorical features.*
3. ***Model Selection:*** *Choose an appropriate model for binary classification.*
4. ***Model Training:*** *Train the model using the "IT" dataset.*
5. ***Prediction:*** *Predict metastatic disease for patients in the "EV" dataset and return both the predicted labels and probabilities in a CSV file.*
6. ***Exporting Results:*** *Save the results in a CSV file and provide it to you.*

*Let's start by loading the data and inspecting its structure.*

*It appears that the first row contains the column names, and the subsequent rows contain the data. I'll clean up the data by setting the first row as the header and then continue with the data preprocessing steps.*

*Let's proceed with setting the appropriate headers and filtering the dataset based on the "IT" and "EV" labels to separate the training and test sets.*

*The training data has been successfully separated.*

*Next, I will proceed with data preprocessing:*

1. *Handle missing values.*
2. *Standardize or normalize numerical features.*

3. Encode categorical features.

*Let's first check for missing values in the training dataset.*

*There are missing values in some of the columns of the training dataset:*

1. Age at diagnosis of the first tumor: 1 missing value
2. Plasma NMN pg/ml: 1 missing value
3. Plasma MN pg/ml: 6 missing values
4. Plasma MTY pg/ml: 1 missing value
5. Spherical volume of the primary tumor: 21 missing values

*For numerical features, I will impute the missing values with the median of the respective columns, as the median is less sensitive to outliers than the mean. For categorical features, I would typically use the mode, but it seems there are no missing values in the categorical columns.*

*Next, I will:*

1. Impute the missing values for the columns with medians.
2. Standardize the numerical features.
3. Encode the categorical features.

*Let's proceed with these preprocessing steps.*

*The training data has been successfully preprocessed.*

*Next, I will:*

1. Separate the features (X) from the target variable (y).
2. Train a model using the training data. For this task, I'll initially use a Gradient Boosting Classifier, which is a powerful ensemble method suitable for binary classification problems.
3. After training, I will preprocess the test set similarly to the training set.
4. Predict the metastatic disease for patients in the test set and save the predictions in a CSV file.

*Let's start by separating the features and the target variable and training the model.*

*The test dataset has been successfully preprocessed.*

*Next, I will:*

1. Predict metastatic disease for the patients in the test set.
2. Save these predictions and the associated probabilities in a CSV file.

*Let's proceed with the predictions.*

*The predictions for metastatic disease in the test set patients, along with the associated probabilities, have been successfully saved in a CSV file.*

*You can **download the predictions here**.*

*Please use this file to compute the evaluation metrics (AUC, accuracy, F1 score, sensitivity, and specificity) using the ground truth labels of the test data. If you have any further questions or need assistance with anything else, please let me know!"*

**User:** *"Can you give me the parameters of the model that you chose?*
*such as maximum depth of the individual decision trees, number of trees, minimum samples per leaf, minimum samples per split, learning rate or loss function?*
*- Did you do grid search optimization?*
*- Did you have validation data?*
*- Did you do any regularization?"*

**ChatGPT ADA:** *"Certainly! Let's address each of your questions:*

1. **Parameters of the Model:**
   - *I used the default parameters for the Gradient Boosting Classifier in this initial run. I'll list them out for you.*
2. **Grid Search Optimization:**
   - *No, I did not perform grid search optimization in this initial approach. Grid search or other hyperparameter optimization techniques can be employed to potentially improve the model's performance.*
3. **Validation Data:**
   - *I did not explicitly set aside a validation set in this approach. Typically, a portion of the training data would be used as a validation set to fine-tune the model, especially during hyperparameter optimization. This helps to ensure that the model generalizes well and doesn't overfit to the training data.*
4. **Regularization:**
   - *Gradient Boosting inherently has some form of regularization due to the additive nature of boosting, which builds one tree at a time. However, further regularization can be introduced by adjusting hyperparameters, such as the learning rate or by using shrinkage.*

*Now, let's retrieve the parameters of the Gradient Boosting Classifier that was used.*

*I apologize for the inconvenience. It seems there was an unexpected issue accessing the model parameters.*

*However, I can provide you with the default parameters for the Gradient Boosting Classifier:*

1. **Maximum Depth of Individual Trees (max_depth):** *By default, this is set to 3. It determines how deep each tree can grow during any boosting round.*
2. **Number of Trees (n_estimators):** *By default, it's set to 100. This is the number of boosting stages or trees to be run.*
3. **Minimum Samples per Leaf (min_samples_leaf):** *By default, it's set to 1. This parameter specifies the minimum number of samples required to be at a leaf node.*
4. **Minimum Samples per Split (min_samples_split):** *By default, it's set to 2. This parameter specifies the minimum number of samples required to split an internal node.*

5. **Learning Rate (learning_rate):** By default, it's set to 0.1. It shrinks the contribution of each tree, adding a sort of regularization. Lower values can result in better generalization but would require increasing the number of trees ('**n_estimators**').
   6. **Loss Function (loss):** By default, it's set to 'deviance' for classification, which refers to logistic regression for binary classification.

*For better model performance, I would typically perform grid search optimization using a validation set to fine-tune these hyperparameters. Additionally, adjusting the learning rate and '**n_estimators**' in tandem can introduce regularization, helping to prevent overfitting.*

*Would you like me to perform hyperparameter tuning using grid search and a validation set on this data?"*

**User:** *"No, I don't want new calculations. I'm jus asking what you did there. Please report what you used"*

**ChatGPT ADA:** *"Certainly, I apologize for the oversight. Here are the parameters I used based on the default settings of the Gradient Boosting Classifier:*

   1. **Maximum Depth of Individual Trees (max_depth):** *3*
      - *This determines how deep each tree can grow during any boosting round.*
   2. **Number of Trees (n_estimators):** *100*
      - *This specifies the number of boosting stages or trees to be run.*
   3. **Minimum Samples per Leaf (min_samples_leaf):** *1*
      - *Specifies the minimum number of samples required to be at a leaf node.*
   4. **Minimum Samples per Split (min_samples_split):** *2*
      - *Specifies the minimum number of samples required to split an internal node.*
   5. **Learning Rate (learning_rate):** *0.1*
      - *Shrinks the contribution of each tree, adding a form of regularization.*
   6. **Loss Function (loss):** *'deviance'*
      - *Refers to logistic regression for binary classification.*

*These parameters are the default settings of the Gradient Boosting Classifier in scikit-learn. I did not adjust or fine-tune them for the initial model run."*

**Supplementary File S2: Complete dialogue with ChatGPT Advanced Data Analysis (ADA) to predict carcinoma of the distal oesophagus and oesophagogastric junction**. To facilitate reading and differentiating between prompts and responses, prompts are underlined.

**User [attachment: "train.csv"]:** *"Oesophageal squamous cell carcinoma and adenocarcinoma of the oesophagogastric junction have a dismal prognosis, and early detection is key to reducing mortality. However, early detection depends on upper gastrointestinal endoscopy, which is not feasible to implement at a population level. We aim to develop and validate a fully automated machine learning-based prediction tool integrating a minimally invasive sponge cytology test and epidemiological risk factors for screening of oesophageal squamous cell carcinoma and adenocarcinoma of the oesophagogastric junction before endoscopy.*

*Use "train" as the training set and "test_unlabeled" as the test set.*
*As they are in different files, first I will upload the train set. Then a file for an explanation of the features. And finally, I will upload the test set in the 3rd message.*
*Use your best model, we want to get the most accurate results.*
*Our target is "GroundTruth_bi", i.e., with high-grade lesions or without high-grade lesions, which is removed in the test file.*
*Use your best model please. We want the most accurate results.*

*Note that the test set is unlabeled. Please return your predictions for individual patients in a csv file for me. As I have access to the ground truth labels of test data, I would like to calculate the evaluation metrics (AUC, accuracy, F1 score, sensitivity, and specificity) myself. Therefore, in addition to the predicted label for each patient, also return the probability of the prediction to be used for the calculation of AUC.*

*Please be careful in choosing your threshold. We want the most accurate results in terms of all the metrics. Choose a threshold that maximizes the sum of sensitivity and specificity, while getting the best accuracy."*

**ChatGPT ADA:** *"Thank you for providing the training dataset. Before moving forward, I will:*

1. *Load the dataset and perform an initial inspection to understand its structure and content.*
2. *Investigate the distribution of the target variable and evaluate the features for any missing values or anomalies.*

*Let's start by loading and inspecting the "train.csv" file.*

*The training dataset contains 7,899 entries with 150 columns. The columns represent various features, including the target variable "GroundTruth_bi".*

*Next, I'll check the distribution of the target variable "GroundTruth_bi" and examine if there are any missing values in the dataset. This will help understand the balance of the classes and the completeness of the data.*

*The target variable "GroundTruth_bi" has the following distribution:*

- *Class 0 (without high-grade lesions): Approximately 97.46%*
- *Class 1 (with high-grade lesions): Approximately 2.54%*

*This indicates a significant class imbalance, which we need to take into account during modeling. Additionally, there are no missing values in the dataset, which is great.*

*Next, please upload the file containing explanations of the features, and subsequently, the test dataset, so we can continue our analysis and modeling.*

*Please proceed to upload the file containing explanations of the features. Once we have that information, you can then upload the test dataset."*

**User [attachment: "Feature explanations.xlsx"]**

**ChatGPT ADA:** *"Thank you for providing the feature explanations file. I'll first load and inspect this file to understand the details of the features.*

*The feature explanations file provides a mapping between the feature labels and their corresponding full names. This will be helpful for reference as we proceed with the analysis.*

*Now, please upload the test dataset ("test_unlabeled") so that we can continue with the modeling process."*

**User [attachment: "test_unlabeled.csv"]**

**ChatGPT ADA:** *"Thank you for providing the test dataset. Let's proceed with the following steps:*

1. *Load and inspect the test dataset.*
2. *Preprocess the training and test datasets for modeling.*
3. *Train a machine learning model on the training dataset.*
4. *Optimize the threshold for prediction.*
5. *Make predictions on the test dataset.*

*First, I will load and inspect the "test_unlabeled.csv" file.*

*The test dataset contains 6,698 entries with 172 columns. Interestingly, the test dataset has more columns than the training dataset, which suggests there might be additional features present in the test dataset that are not in the training dataset.*

*To address this discrepancy, I will:*

1. *Identify and drop the columns in the test dataset that are not present in the training dataset.*
2. *Ensure that the training and test datasets are aligned in terms of features.*
3. *Preprocess the datasets for modeling.*

*Let's start by aligning the feature sets of the training and test datasets.*

*The test dataset is now aligned with the training dataset in terms of features.*
*Next, I'll preprocess the data for modeling:*

1. *Separate the features and the target in the training dataset.*
2. *Scale the features to ensure they have the same scale, which will help in the modeling process.*

*Following this, I'll train a machine learning model using the training dataset. Given the class imbalance and the importance of accurate predictions, I will use the Gradient Boosting Classifier, a robust algorithm that often performs well in such scenarios. Let's proceed with the preprocessing and training.*

*It appears there's a ValueError, which is likely due to the presence of non-numeric data in the dataset. I'll identify the non-numeric columns and decide on the best action (e.g., encoding or dropping) for each of them.*

*Let's identify and inspect the non-numeric columns in the training data.*

*We have identified two non-numeric columns:*

1. ***'No.'***: *This seems like an identifier or a unique code for each entry. Typically, identifiers don't provide meaningful information for predictive modeling, so we'll drop this column.*
2. ***'Ethnic'***: *This is a categorical variable indicating ethnicity. We'll encode this variable to make it suitable for modeling.*

*I'll proceed with preprocessing by dropping the **'No.'** column and encoding the **'Ethnic'** column. After that, we'll continue with scaling and training the model.*

*It seems I missed dropping the 'No.' column from the training dataset, which resulted in the error. I will correct this oversight and continue with preprocessing and training the model.*

*I apologize for the oversight. It appears there's another non-numeric column, **'ESOXB112'**, causing an error during scaling. I'll inspect this column and make the necessary adjustments to proceed with modeling. Let's take a look at the unique values of this column in the training data.*

*It seems I made an error in referencing the column name. The column name **'ESOXB112'** does not exist in the dataset. The error message from the scaler suggests that there's another non-numeric column we need to address.*

*Let's identify and inspect all non-numeric columns in the dataset again to ensure we have addressed all such columns.*

*I apologize for the confusion. It appears that the columns we're seeing are the result of one-hot encoding the **'Ethnic'** column, and they are already in numeric format. Thus, the error we encountered is likely due to another issue.*

*I'll further investigate the training and test datasets to identify any discrepancies or columns that could be causing the scaling error.*

*The issue has been identified. The **'CheckID'** column in the test dataset is of type **'object'**, while it was of type **'float64'** in the training dataset. This discrepancy is causing the error during scaling.*

*To address this, I will convert the **'CheckID'** column in the test dataset to a numeric type. Let's proceed with this correction and continue with modeling.*

*I apologize for the oversight. It seems that due to the previous errors, the Gradient Boosting Classifier (**'clf'**) was not defined in this scope.*

*I'll redefine it and proceed with training."*

**User:** "Please continue. Do your best in getting the most accurate results."

**ChatGPT ADA:** *"Of course! I'll continue by redefining the Gradient Boosting Classifier, training it on the processed training data, and then making predictions on the test dataset.*

*Given the significant class imbalance, I'll also set the **'class_weight'** parameter to help the classifier handle the imbalance better. Let's proceed.*

*The Gradient Boosting Classifier achieved a training accuracy of approximately 99.48%. While this is a good sign, we need to be cautious about potential overfitting.*

*Next, I'll make predictions on the test dataset. To optimize the threshold for prediction, I'll use the training data to find a threshold that maximizes the sum of sensitivity and specificity, while still*

*achieving good accuracy. This approach will help us achieve the best results in terms of all the metrics you specified.*

*Let's proceed with this optimization.*

*The optimal threshold, which maximizes the sum of sensitivity and specificity while ensuring good accuracy on the training data, is approximately 0.1. Using this threshold, we achieve an accuracy of about 98.59%.*

*Next, I'll use this threshold to make predictions on the test dataset and provide you with the predicted labels and their corresponding probabilities. Let's proceed.*

*It seems there's an issue with the test dataset containing NaN values after converting the '**CheckID**' column to a numeric type. This likely happened because some values in the '**CheckID**' column couldn't be converted to numbers, resulting in NaNs.*

*To address this, I'll fill any NaN values in the test dataset with a placeholder value (e.g., the median of the column) and then proceed with making predictions. Let's make this correction and continue.*

*I've successfully made predictions on the test dataset using the optimized threshold.*

*You can **download the predicted results here**.*

*This CSV file contains the '**PatientID**', the '**PredictedLabel**' (0 or 1), and the '**Probability**' of the prediction being 1 (indicating the presence of high-grade lesions).*

*Please use this file to evaluate the predictions against the ground truth labels. If you have any further questions or need additional assistance, please let me know!"*

**User:** "can you give me the parameters of your model that you chose?
such as maximum depth of the individual decision trees, number of trees, minimum samples per leaf, minimum samples per split, learning rate or loss function?"

**ChatGPT ADA:** "*Certainly! I used the default parameters for the Gradient Boosting Classifier, as I aimed to provide an initial model. For a more robust and optimized model, hyperparameter tuning (e.g., using GridSearchCV or RandomizedSearchCV) would be recommended.*

*Here are the parameters of the Gradient Boosting Classifier model I used:*

- ***Number of Trees (n_estimators):** The number of boosting stages (or trees) to be run.*
- ***Learning Rate (learning_rate):** This determines the contribution of each tree to the final outcome. Lower values typically require more trees but can result in better generalization.*

- *Maximum Depth (max_depth): The maximum depth of the individual regression estimators (trees).*
- *Minimum Samples per Leaf (min_samples_leaf): The minimum number of samples required to be at a leaf node.*
- *Minimum Samples per Split (min_samples_split): The minimum number of samples required to split an internal node.*
- *Loss Function (loss): The loss function to be optimized.*

*Let's fetch these parameters from the trained model.*

*I apologize for the error. Let me retrieve the parameters of the trained Gradient Boosting Classifier model for you.*

*I apologize for the inconvenience. Let me try retrieving the parameters again.*

*I apologize for the continued inconvenience. It seems there's a technical issue preventing me from displaying the parameters directly.*

*However, I can still provide you with the default values for the Gradient Boosting Classifier, which I used:*

- *Number of Trees (n_estimators): 100*
- *Learning Rate (learning_rate): 0.1*
- *Maximum Depth (max_depth): 3*
- *Minimum Samples per Leaf (min_samples_leaf): 1*
- *Minimum Samples per Split (min_samples_split): 2*
- *Loss Function (loss): 'deviance' (which refers to logistic regression for binary classification)*

*If you have further questions or if there's anything else I can assist you with, please let me know!"*

**User:** "did you do grid search?"

**ChatGPT ADA:** "*No, I did not perform a grid search in this instance. I utilized the default parameters of the Gradient Boosting Classifier to provide an initial model.*

*For a more robust and optimized model, hyperparameter tuning through methods such as GridSearchCV or RandomizedSearchCV is indeed recommended. This can help identify the best combination of parameters that would maximize the performance of the model on a given dataset.*

*Would you like me to perform grid search-based hyperparameter tuning for the Gradient Boosting Classifier on your dataset?"*

**User:** "No, I don't want new calculations. I'm jus asking what you did there. How much of data did you use for validation?"

**ChatGPT ADA:** "*Understood.*

*In the steps I followed, I did not explicitly set aside a separate validation set. Instead, I used the entire training dataset to train the Gradient Boosting Classifier. The optimization of the threshold was also performed on the training data.*

*Typically, it's a good practice to set aside a portion of the training data (e.g., 20% to 30%) as a validation set to evaluate the model's performance on unseen data and to optimize hyperparameters, such as the decision threshold. This helps in reducing the risk of overfitting and gives a better idea of the model's potential performance on new, unseen data.*

*If you have any other questions or if there's anything else I can clarify, please let me know!"*

**User:** "Did you use Friedman MSE?"

**ChatGPT ADA:** "*By default, the Gradient Boosting Classifier in scikit-learn uses the '**friedman_mse**' criterion to measure the quality of a split. The '**friedman_mse**' is Friedman's mean squared error, which is a modification of the classic mean squared error and is generally considered to be a better criterion for gradient boosting.*

*So, yes, in the model I trained, the '**friedman_mse**' criterion was used since I relied on the default settings of the Gradient Boosting Classifier.*

*If you have more questions or need further information, please let me know!"*

**User:** "Did you do any regularization?"

**ChatGPT ADA:** "*For the Gradient Boosting Classifier model I trained, I used the default regularization settings provided by scikit-learn.*

*Gradient Boosting inherently includes some form of regularization through the following mechanisms:*

1. ***Shrinkage (or Learning Rate):*** *By default, the learning rate ('**learning_rate**') is set to 0.1 in scikit-learn's Gradient Boosting Classifier. This is a type of regularization where each tree added to the ensemble is shrunk by this factor. A smaller learning rate means the model is more robust to overfitting, but it typically requires more trees to get good performance.*

2. ***Stochastic Gradient Boosting:*** *This introduces randomness into the algorithm by subsampling the training data at each boosting iteration. The fraction of training data used in each iteration is controlled by the '**subsample**' parameter, which is set to 1.0 by default in scikit-learn (meaning it uses all the data).*

3. ***Tree Constraints:*** *The depth of the trees ('**max_depth**'), minimum samples per leaf ('**min_samples_leaf**'), and minimum samples required to make a split ('**min_samples_split**') act as regularization parameters. In the model I trained, I used the default values: '**max_depth=3**', '**min_samples_leaf=1**', and '**min_samples_split=2**'.*

*However, I did not explicitly set or tune regularization hyperparameters like '**alpha**' (L1 regularization) or '**lambda**' (L2 regularization), which are more common in algorithms like XGBoost.*

*If you're interested in a more regularized model, or if you'd like more details on any other aspect, please let me know!"*

**User:** "Why did you choose logistic regression as the loss function and not the Log loss?"

**ChatGPT ADA:** "*In scikit-learn's Gradient Boosting Classifier, the parameter '**loss**' specifies the loss function to be optimized. The option "**deviance**" corresponds to logistic regression, which, for binary classification problems, is equivalent to optimizing the log loss. Specifically:*

- *"**deviance**": This is used for logistic regression and is equivalent to the log loss for binary classification problems. It's the negative log-likelihood of the true labels given the predicted probabilities.*

*To clarify, when I mentioned "deviance" as the loss function, it is essentially the log loss for binary classification in the context of scikit-learn's Gradient Boosting Classifier. The naming is a bit specific to the scikit-learn library. In other contexts or libraries, this might be more explicitly referred to as log loss."*

**Supplementary File S3: Complete dialogue with ChatGPT Advanced Data Analysis (ADA) to predict hereditary hearing loss**. To facilitate reading and differentiating between prompts and responses, prompts are underlined.

**User [attachment: "Discovery Set.xlsx"]:** *"Hereditary hearing loss (HHL) is the most common sensory deficit, which highly afflicts humans. With gene sequencing technology development, more variants will be identified and support genetic diagnoses, which is difficult for human experts to diagnose. This study aims to develop a genetic diagnosis model of HHL-related variants of GJB2, SLC26A4 and MT-RNR1.*
*Please perform the prediction for me.*
*We want the most accurate results.*

*Use "Discovery Set" as the training set and "Validation Set_unlabeled" as the test set. As they are in different files, first I will upload the train set. And I will upload the test set in the 2nd message.*

*Note that the test set is unlabeled. Please return your predictions for individual patients in a csv file for me. As I have access to the ground truth labels of test data, I would like to calculate the evaluation metrics (AUC, accuracy, F1 score, sensitivity, and specificity) myself. Therefore, in addition to the predicted label for each patient, also return the probability of the prediction to be used for the calculation of AUC."*

**ChatGPT ADA:** *"Thank you for providing the "Discovery Set" file. Please go ahead and upload the "Validation Set_unlabeled" so we can proceed with the prediction process."*

**User [attachment: "Validation Set_unlabeled.xlsx"]**

**ChatGPT ADA:** *"Great! Let's start by examining the contents of both files to understand the structure of the data and determine the features and target variable.*

*I'll first inspect the first few rows of the "Discovery Set" to get an overview of its structure.*

*The "Discovery Set" seems to have the gene variants as column names and binary values (0 or 1) indicating the presence or absence of these variants for each patient. The first column appears to be the target variable, labeled "Diagnoses".*

*Next, I'll inspect the first few rows of the "Validation Set_unlabeled" to understand its structure.*

*The "Validation Set_unlabeled" has a similar structure to the "Discovery Set", with the gene variants as column names and binary values (0 or 1) indicating the presence or absence of these variants for each patient. Since this set is unlabeled, it lacks the "Diagnoses" column, which is expected.*

*To proceed:*

1. *I'll set the appropriate column headers for both datasets using the first row and then remove the first row.*
2. *I'll separate the features and target variable from the "Discovery Set".*
3. *I'll train a predictive model on the "Discovery Set".*
4. *I'll use the trained model to make predictions on the "Validation Set_unlabeled" and save the predicted labels and probabilities in a CSV file for you.*

*Let's start with step 1: Setting appropriate column headers and cleaning up the data.*

*It seems there are some cells in the "Discovery Set" that contain whitespace or non-numeric values, which caused an error when converting the data to numeric format.*

*I'll handle these values by filling them with zeros (assuming the absence of a gene variant) and then attempt the conversion again.*

*The "Discovery Set" has been successfully cleaned and converted to a numeric format.*

*Next, I'll:*

1. *Separate the features and target variable from the "Discovery Set".*
2. *Train a predictive model on the "Discovery Set" using a robust classifier that should provide good accuracy. Given the nature of the data (binary features, potential class imbalance), I'll start with a Random Forest classifier, which tends to perform well for such tasks.*
3. *Validate the model's performance using cross-validation on the training data to get an idea of its expected accuracy.*
4. *Use the trained model to make predictions on the "Validation Set_unlabeled".*

*Let's begin by separating the features and target variable, and then training the Random Forest classifier.*

*The Random Forest classifier achieved an average accuracy of approximately 74.53%74.53% on the "Discovery Set" using 5-fold cross-validation. This gives us a preliminary indication of the model's performance.*

*Next, I will use the trained model to make predictions on the "Validation Set_unlabeled". I'll save the predicted labels and their associated probabilities in a CSV file for you.*

*I've made predictions on the "Validation Set_unlabeled" and saved the results in a CSV file. You can **download the predicted results here**.*

*The CSV contains two columns:*

1. *"Predicted Label": The predicted diagnosis for each patient.*

2. "Probability": The probability associated with the prediction, which can be used to calculate the AUC.

*Please let me know if you need any further assistance!"*

**User:** *"Can you give me the parameters of the model that you chose?*
*Such as maximum depth of the individual decision trees, number of trees, minimum samples per leaf, minimum samples per split, learning rate, or loss function?*
*- Did you do grid search optimization?*
*- Did you have validation data?*
*- Did you do any regularization?"*

**ChatGPT ADA:** *"Certainly! Here are the details of the model I used:*

1. **Model Type:** *Random Forest Classifier*
2. **Number of Trees (n_estimators):** *100 (this is the default value)*
3. **Maximum Depth of the Trees (max_depth):** *None (trees are expanded until they contain less than the '**min_samples_split**' samples; this is the default behavior)*
4. **Minimum Samples per Leaf (min_samples_leaf):** *1 (default value)*
5. **Minimum Samples per Split (min_samples_split):** *2 (default value)*
6. **Learning Rate:** *Not applicable to Random Forest. This parameter is typically associated with gradient boosting models.*
7. **Loss Function:** *Not applicable to Random Forest. This parameter is typically associated with models like logistic regression or gradient boosting.*

*Answers to your other questions:*

- **Grid Search Optimization:** *No, I did not perform grid search optimization in this instance. I used the default parameters of the Random Forest classifier from scikit-learn. Grid search can be performed to fine-tune the model and possibly achieve better results, but it can be computationally expensive.*
- **Validation Data:** *I used cross-validation on the "Discovery Set" to validate the model's performance. Cross-validation provides a good estimate of the model's expected performance on unseen data by partitioning the training data into multiple subsets and evaluating the model on each one.*
- **Regularization:** *Random Forest inherently provides a form of regularization by averaging predictions across multiple decision trees. There's no specific regularization parameter like there is in models like logistic regression or Lasso. However, parameters like '**max_depth**', '**min_samples_leaf**', and '**min_samples_split**' can act as regularization controls by influencing the complexity of the individual trees.*

*If you'd like to further optimize the model using grid search or explore other regularization techniques, please let me know!"*

**User:** *"No, I don't want new calculations. I'm just asking what you did there. Please report what you used"*

**ChatGPT ADA:** *"Certainly, I apologize for the oversight. Here's a concise summary of the steps and parameters I used:*

1. ***Model Type****: Random Forest Classifier*

2. ***Parameters:***

    - ***Number of Trees (n_estimators):*** *100*
    - ***Maximum Depth of the Trees (max_depth):*** *Not set (trees can grow until they contain less than the min_samples_split samples)*
    - ***Minimum Samples per Leaf (min_samples_leaf):*** *1*
    - ***Minimum Samples per Split (min_samples_split):*** *2*

3. ***Model Validation:*** *I validated the model's performance using 5-fold cross-validation on the "Discovery Set". This method partitions the training data into 5 subsets, trains the model on 4 of them, and validates on the remaining one. This process is repeated 5 times, with each subset serving as the validation set once.*

4. ***Grid Search Optimization:*** *I did not perform grid search optimization. The model was trained using the default parameters of the Random Forest classifier from scikit-learn.*

5. ***Regularization:*** *The Random Forest algorithm inherently provides a form of regularization by averaging predictions across multiple decision trees. No additional regularization was applied.*

*I hope this summary provides clarity. Let me know if you need any more information!"*

**User:** *"Could you please explain why did you handled the missing values by replacing them with zeors and not with median of training?"*

**ChatGPT ADA:** *"Certainly. Handling missing values is an important step in the data preprocessing pipeline, and the choice of method often depends on the nature of the data and the problem at hand.*

*Here's why I chose to replace the missing values with zeros in this case:*

1. ***Nature of the Data:*** *The data represents the presence (1) or absence (0) of certain genetic variants in patients. Given this binary representation, using the median (which would* be either 0 or 1) might introduce bias. For instance, if a certain genetic variant is rare in the population and is missing in a sample, replacing it with a '1' (if the median is 1) might give a false positive for that variant.*

2. ***Semantics:** In genetic studies, when a particular variant is not detected or is missing, it is often treated as absent. Replacing with zero aligns with this interpretation.*

3. ***Sparse Data:** Genetic datasets often have a sparse nature, with many variants being absent (represented by zeros) for many individuals. Replacing missing values with zeros respects this sparsity.*

*That said, replacing with the median (or another imputation method) could be appropriate in different contexts or for other types of data. For instance, in datasets with continuous features, median imputation is a common choice as it doesn't introduce bias the way a mean imputation might in the presence of outliers.*

*In this case, given the nature and context of the data, I believed that replacing with zeros was the most appropriate choice. However, it's worth noting that different imputation strategies can be experimented with, and the choice should be based on domain knowledge, the nature of the data, and empirical results."*

**User:** *"- Did you use bootstrapping?*
*- What was 'max_features'?"*

**ChatGPT ADA:** *"In the Random Forest implementation I used, here are the details regarding bootstrapping and max_features:*

1. ***Bootstrapping:** Yes, bootstrapping was used. By default, the Random Forest classifier in scikit-learn uses bootstrapping when sampling data for building each tree. This means that each tree in the forest is built on a bootstrap sample (a sample drawn with replacement) from the training data.*

2. ***max_features:** I used the default value set by scikit-learn for classification tasks, which is '**max_features=sqrt(n_features)**'. For a classification problem, this means that the square root of the total number of features is the maximum number of features considered for splitting at each node.*

*Would you like any additional information or details on these parameters?"*

**Supplementary File S4: Complete dialogue with ChatGPT Advanced Data Analysis (ADA) to predict cardiac amyloidosis**. To facilitate reading and differentiating between prompts and responses, prompts are underlined.

**User [attachment: "training_data.csv"]:** *"The goal here is to identify patients at risk for wild-type transthyretin amyloid cardiomyopathy ("cohort_flag" column).*

*Use "training_data" as the training set and "test_data_unlabeled" as the test set. As they are in different files, first I will upload the train set. And I will upload the test set in the 2nd message.*

*Please use your best model, we want to get the most accurate results.*

*Note that the test set is unlabeled. Please return your predictions for individual patients in a csv file for me. As I have access to the ground truth labels of test data, I would like to calculate the evaluation metrics (AUC, accuracy, F1 score, sensitivity, and specificity) myself. Therefore, in addition to the predicted label for each patient, also return the probability of the prediction to be used for the calculation of AUC."*

**ChatGPT ADA:** *"Alright, let's start by loading and examining the training data to understand its structure and content.*

*The training data consists of several columns. The key columns are:*

1. *'patient_id': A unique identifier for each patient.*
2. *'cohort_type': It seems to indicate the type of the cohort the patient belongs to.*
3. *'cohort_flag': Our target variable, where '1' indicates the patient is at risk for wild-type transthyretin amyloid cardiomyopathy, and '0' indicates otherwise.*
4. *The remaining columns seem to represent various medical conditions or symptoms, with binary values (0 or 1) indicating the absence or presence of the condition for each patient.*

*To proceed:*

1. *I'll separate the features from the target variable in the training data.*
2. *I'll preprocess the data, if necessary.*
3. *I'll train a model on this data.*

*Before that, please upload the test data so that we can keep the flow seamless."*

**User [attachment: "test_data_unlabeled.csv"]**

**ChatGPT ADA:** *"Let's start by loading and inspecting the test data to ensure it's consistent with the training data.*

*The test data seems consistent with the training data in terms of features.*

*Next steps:*

1. *Separate features from the target variable in the training data.*

2. Preprocess the data (if necessary).
3. Train a model using the training data.
4. Make predictions on the test data.

*Let's begin by separating the features and the target in the training data.*

*The training set has 1,712 samples with 1,874 features each, and the test set has 430 samples with the same number of features.*

*Given the high dimensionality of the data, I'll use a Gradient Boosting model (specifically, the LightGBM implementation) for this task. LightGBM can handle large datasets with many features and is known for its speed and efficiency. Additionally, it typically provides competitive performance in terms of accuracy.*

*Let's train the model on the training data.*

*Apologies, it seems that the LightGBM library is not available in this environment.*

*Instead, I'll use another powerful ensemble method called Random Forest, which is also suitable for high-dimensional data. Let's proceed with training a Random Forest classifier.*

*The Random Forest classifier has produced probabilities for each patient in the test set.*

*Next, I'll threshold these probabilities at 0.5 to make binary predictions:*

- *If the probability is ≥ 0.5, the predicted label is 1.*
- *If the probability is < 0.5, the predicted label is 0.*

*Afterward, I'll prepare the CSV file with the patient IDs, predicted labels, and associated probabilities as requested.*

*I've prepared the CSV file with the patient IDs, predicted labels, and associated probabilities.*

*You can **download the predicted results here**.*

*Once you've downloaded the file, you can use the true labels from the test set to calculate the evaluation metrics as you mentioned earlier. If you have any questions or need further assistance, please let me know!"*

**User:** *"Can you give me the parameters of the model that you chose?*
*Such as maximum depth of the individual decision trees, number of trees, minimum samples per leaf, minimum samples per split, learning rate, or loss function?*
*- Did you do grid search optimization?*
*- Did you have validation data?*
*- Did you do any regularization?"*

**ChatGPT ADA:** "*Certainly! Here are the details:*

***Random Forest Classifier Parameters:***

1. **Number of Trees (n_estimators):** 1000
   - This is the number of trees in the forest.
2. **Maximum Depth (max_depth):** Not set (trees are expanded until all leaves are pure or until all leaves contain less than '**min_samples_split**' samples).
   - The maximum depth of the tree. If not set, nodes are expanded until all leaves contain less than '**min_samples_split**' samples.
3. **Minimum Samples per Leaf (min_samples_leaf):** Default (1)
   - The minimum number of samples required to be at a leaf node.
4. **Minimum Samples per Split (min_samples_split):** Default (2)
   - The minimum number of samples required to split an internal node.
5. **Learning Rate:** Not applicable for Random Forest.
6. **Loss Function:** Not applicable for Random Forest.

*Additional Details:*

- ***Grid Search Optimization:*** *I did not perform grid search optimization in this instance due to time constraints and the computational cost associated with searching over a high-dimensional feature space. However, in a real-world scenario, grid search or random search is often beneficial to fine-tune hyperparameters.*

- ***Validation Data:*** *I did not explicitly split the training data into a separate validation set in this instance. Typically, a validation set or cross-validation is used to tune hyperparameters and prevent overfitting. If we wanted to fine-tune the model's hyperparameters, we would split the data into training and validation sets or use techniques like k-fold cross-validation.*

- ***Regularization:*** *The Random Forest algorithm inherently incorporates a form of regularization by averaging predictions from a large number of decorrelated trees. Explicit regularization terms (like L1 or L2 regularization) aren't used in Random Forests in the same way they might be in linear models or gradient boosting models. However, parameters like '**max_depth**', '**min_samples_leaf**', and '**min_samples_split**' can act as regularization parameters by constraining the complexity of the individual trees.*

*I hope this provides clarity on the model and the choices made. If you have more questions or require adjustments, please let me know!"*

**User:** *"Did you do data scaling before training?"*

**ChatGPT ADA:** "*No, I did not scale the data before training. Here's why:*

*For tree-based algorithms like Random Forest, data scaling is generally not required. This is because decision trees split the data based on feature thresholds and are not sensitive to the scale of the data. The relative magnitude of features doesn't influence the performance of tree-based models in the same way it might for distance-based algorithms (e.g., k-NN or SVM) or gradient-based algorithms (e.g., logistic regression or neural networks).*

*However, if we were to use algorithms that are sensitive to feature scales, it would be essential to scale the data before training to ensure optimal performance.*

*Would you like to explore or make any modifications to the current process?"*